\documentclass[conference]{llncs}

\input epsf
\usepackage{graphicx}
\usepackage[margin=0.7in]{geometry}
\usepackage{amsmath}
\usepackage{algorithm}
\usepackage{amsfonts}
\usepackage{amssymb}
\usepackage{caption}
\usepackage{booktabs}
\usepackage{algpseudocode}
\usepackage{tikz}
\usepackage{authblk}
\usepackage{placeins}

\begin{document}

\title{\LARGE Enhancing ASL Recognition with GCNs and Successive Residual Connections}



\author{Ushnish Sarkar\inst{1} \and
Archisman Chakraborti \inst{2}\ \and
Tapas Samanta\inst{1} \and Sarbajit Pal\inst{1} \and Amitabha Das \inst{3}}
\authorrunning{F. Author et al.}
%
\institute{Variable Energy Cyclotron Centre, Kolkata \\
\email{u.sarkar@vecc.gov.in}\and
Harish Chandra Research Institute, Prayagraj
\\
\email{archismanchakraborti@hri.res.in}
 \and
Jadavpur University, Kolkata\\
}
\authorrunning{Sarkar et al.}


\maketitle

\begin{abstract}
This study presents a novel approach for enhancing American Sign Language (ASL) recognition using Graph Convolutional Networks (GCNs) integrated with successive residual connections. The proposed method leverages the MediaPipe framework to extract 21 key landmarks from each hand gesture, which are then used to construct graph representations. We introduce a robust preprocessing pipeline that includes translational and scale normalization techniques to ensure consistency across the dataset. The constructed graphs are subsequently fed into a GCN-based neural architecture, which employs residual connections to mitigate gradient-related issues and improve network stability \cite{He2016}. The architecture's performance was rigorously evaluated on the ASL Alphabet dataset \cite{ASLDataset2018}, achieving state-of-the-art results with a validation accuracy of 99.14\%. Extensive experimentation, including a 5-fold cross-validation, demonstrated the model's superior generalization capabilities. The integration of dropout layers and batch normalization further enhanced the model's robustness against overfitting\cite{StateOfTheArt2020}.

\end{abstract}

%

\section*{Introduction}
Sign language recognition is an essential aspect of human-computer interaction, enabling seamless communication between the deaf and hard-of-hearing communities and the broader population.Signing gloss consists of various parameters , hand-shape being the most prominent among them. Hence, hand-shape recognition is an essential step in sign language translation pipeline.   Among various sign languages, American Sign Language (ASL) is widely used, making its recognition crucial for developing assistive technologies, such as real-time translation systems and educational tools.

Traditional approaches to ASL recognition have primarily relied on image processing techniques and conventional machine learning algorithms. These methods, while effective to some extent, often struggle with the inherent complexity and variability of hand gestures, especially when dealing with intricate finger movements and orientations. Convolutional Neural Networks (CNNs) have been widely adopted to address these challenges due to their ability to learn hierarchical features directly from image data. However, CNNs are inherently limited in their ability to capture the non-Euclidean relationships present in the spatial configurations of hand landmarks, as they are designed to operate on regular grid-like data \cite{Tongi2021},
\cite{Dong2015}, 
\cite{Kang2020}.

To overcome these limitations, Graph Convolutional Networks (GCNs) have emerged as a powerful tool for modeling data that can be naturally represented as graphs. Unlike CNNs, GCNs are capable of directly operating on graph-structured data, allowing for a more expressive and efficient modeling of the spatial dependencies between low level features (in our context, hand landmarks) \cite{Kipf2017}, \cite{Bruna2014}, which is crucial for identifying high level features such as shape.  
In the context of ASL recognition, each hand can be represented as a graph where the nodes correspond to the key landmarks of the hand, and the edges represent the spatial relationships between these landmarks. This graph-based representation provides a more natural way to capture the complex geometric relationships inherent in hand gestures.

Moreover, the inclusion of residual connections within the GCN architecture is motivated by the need to address the issues of vanishing and exploding gradients, which are common in deep neural networks \cite{He2016}. 
In this paper, we present a novel ASL recognition framework that leverages GCNs with successive residual connections to achieve state-of-the-art performance. The core of our approach involves the following key steps:
\begin{enumerate}
    \item \textbf{Landmark Extraction:} We utilize the MediaPipe framework to extract 21 key landmarks from each hand gesture. These landmarks form the basis of our graph-based representation.
    \item \textbf{Graph Construction:} The extracted landmarks are represented as a graph, where each node corresponds to a landmark, and edges represent the spatial relationships between them.
    \item \textbf{GCN-Based Recognition:} The constructed graphs are fed into a GCN-based neural architecture with residual connections, which enhances the network's ability to learn complex spatial dependencies while maintaining stable training dynamics.
\end{enumerate}

Our extensive experiments on the ASL Alphabet dataset \cite{ASLDataset2018} demonstrate that the proposed approach significantly outperforms existing methods, achieving a validation accuracy of 99.14\%. This work sets a new benchmark for ASL recognition and provides a robust framework for future research in this domain.

\section*{Related Works}
The field of American Sign Language (ASL) recognition has seen substantial contributions through the use of various deep learning architectures, particularly Graph Convolutional Networks (GCNs), Convolutional Neural Networks (CNNs), and Graph Attention Networks (GATs). Convolutional Neural Networks (CNNs) have been extensively employed due to their powerful feature extraction capabilities. Pigou et al. (2017) used Temporal Residual Networks (TRNs) combined with CNNs for gesture and sign language recognition, which effectively handled both spatial and temporal aspects of the video data \cite{pigou2017gesture}. Similarly, the work by Rao et al. (2018) focused on isolated sign language recognition using deep CNNs, extracting spatial features from each frame of video sequences and managing temporal dependencies using Long Short-Term Memory (LSTM) networks \cite{rao2018deep}.

The introduction of Graph Convolutional Networks (GCNs) provided a new paradigm by focusing on the spatial relationships between body joints, represented as graph nodes. Yan et al. (2018) pioneered the use of Spatial-Temporal GCNs (ST-GCNs) in the ASL domain, applying these models to the ASL Lexicon Video Dataset (ASLLVD). In their work, skeletal data extracted via the OpenPose framework was used to model the dynamic gestures of signing, capturing both spatial structures and temporal dynamics across video frames \cite{yan2018spatial}. Further advancements were made by integrating GCNs with transformers like BERT (Bidirectional Encoder Representations from Transformers). This hybrid approach, as demonstrated by recent studies, involved using GCNs to model spatial dependencies while BERT managed temporal relationships across frames, resulting in improved accuracy on large-scale datasets such as WLASL \cite{pose2020gcn}.

Graph Attention Networks (GATs) were introduced to further enhance the capabilities of GCNs by incorporating attention mechanisms. GATs allowed models to selectively focus on more critical joints within the graph, improving the representation learning process. Although the application of GATs in ASL recognition is relatively new, preliminary research suggests their potential in developing more robust and accurate sign language recognition systems \cite{pose2020gcn}. These methodologies highlight the evolution of ASL recognition from traditional CNN-based methods to more sophisticated graph-based approaches that emphasize the importance of spatial and temporal dynamics.

\section*{Dataset Overview}
The dataset utilized in this study is the ASL Alphabet dataset \cite{nagaraj2018asl}, which consists of images depicting the American Sign Language alphabets. The dataset is organized into 29 classes. These classes include the 26 letters from A to Z and three additional classes for SPACE, DELETE, and NOTHING.

\textbf{Data Splits:}
The dataset was split with a 80-20-20 split. This means that 80 \% of the entire dataset was used for training the network, 20\% of the training split was used for validation during training of the network and 20\% of the dataset was used for testing the network after training.

\vspace{-0.1in}
\section{Proposed Methods}
Our main aim is to optimise validation metrics like accuracy, prescision, recall, F1-score and Area Under the Curve(AUC) score over the American sign Language(ASL) Dataset using a novel convolutional graph based neural architecture and show that it outperforms existing state-of-the art architectures in the above-mentioned dataset. 
In this section, we first describe the pre-processing steps we undertook on the ASL dataset followed by our proposed network architecture.

\begin{figure}
    \centering
    \includegraphics[width=0.4\linewidth]{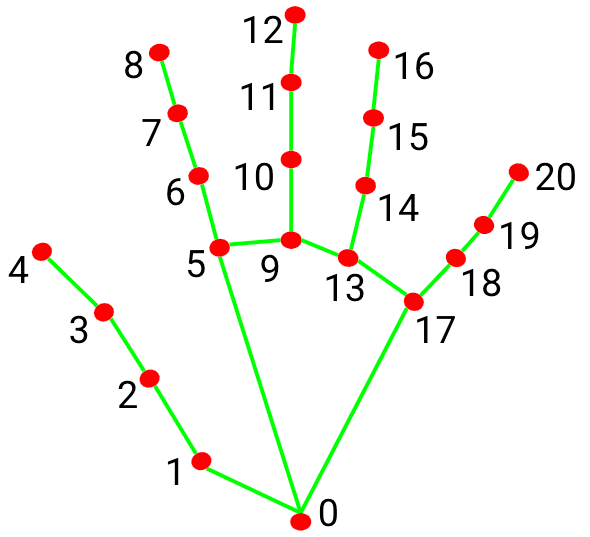}
    \caption{The 21 Hand landmarks detected by MediaPipe hands solutions package.}
    \label{fig:Mediapipe_landmarks}
\end{figure}

\subsection{Preprocessing}
We extracted the required human hand landmarks which were used as graph nodes for recognition using MediaPipe. Following this, we applied Translational and Scale normalizations on the extracted landmarks to maintain consistency across all instances of training data.

\begin{algorithm}
\caption{Preprocessing Pipeline for ASL Recognition}
\begin{algorithmic}[1]
\For{each sample $s$ in $D$}
    \State Extract 21 landmarks: $\mathbf{x}_i \in \mathbb{R}^4$
    \State Calculate corresponding joint angles: $\mathbf{\theta}_i \in [-2 \pi, 2 \pi]$
    \State Translational Normalization: $\mathbf{x}_i' = \mathbf{x}_i - \mathbf{x}_0$
    \State Compute pairwise distances: $d_{ij} = \|\mathbf{x}_i' - \mathbf{x}_j'\|_2$
    \State Maximum distance: $d_{\text{max}} = \max(d_{ij})$
    \State Scale factor: $s = \frac{d_{\text{desired}}}{d_{\text{max}}}$
    \State Scale landmarks: $\mathbf{x}_i'' = s \cdot \mathbf{x}_i'$
    \State Construct $A$ and $E$ matrices for graph $G(V, E)$
\EndFor
\end{algorithmic}
\end{algorithm}

\begin{figure}
    \centering
    \includegraphics[width=0.5\linewidth]{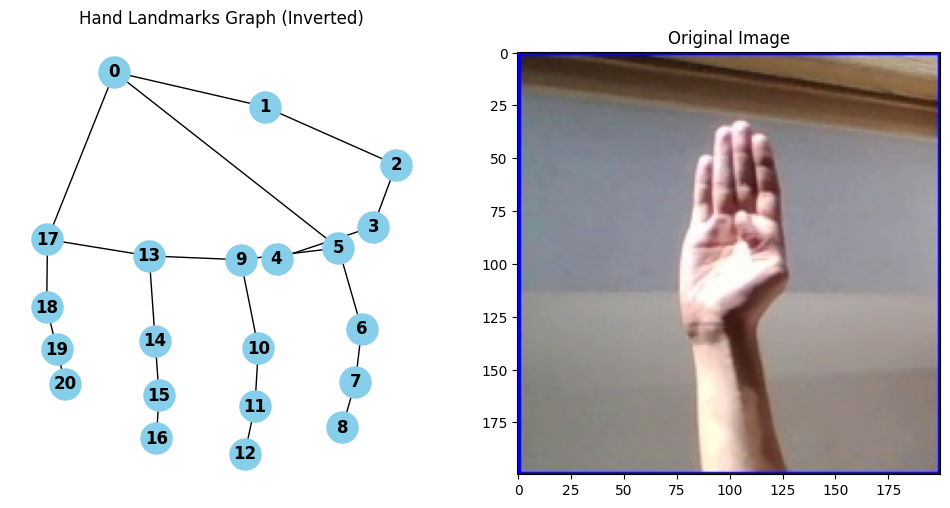}
    \caption{Example of a frame in ASL for the alphabet B and Mediapipe Hand landmark detection on it.}
    \label{fig:Landmarks}
\end{figure}

\subsubsection{MediaPipe Based Graph Construction}
Mediapipe \cite{lugaresi2019mediapipeframeworkbuildingperception} is a cross-platform framework developed by Google for building multimodal applied machine learning pipelines. It offers a variety of pre-trained models and customizable solutions for tasks like hand and face tracking, object detection, and pose estimation.
Mediapipe can detect 21 landmarks per hand. These landmarks correspond to key points on the hand, enabling detailed tracking and analysis of hand movements and gestures. Each landmark is represented by its $x$, $y$ and $z$ Cartesian coordinates.
Figure \ref{fig:Mediapipe_landmarks} shows the extracted 21 landmarks on a sample frame from the ASL dataset.   
For building the graph, we extracted the 21 landmarks as depicted in Figure \ref{fig:Landmarks} and calculated the following angles between various finger joints: 
\begin{itemize}
    \item \textbf{Thumb}: Thumb CMC: $(L_1, L_2, L_3)$; Thumb MCP: $(L_2, L_3, L_4)$
    \item \textbf{Index Finger}: Index Finger MCP: $(L_5, L_6, L_7)$; Index Finger PIP: $(L_6, L_7, L_8)$
    \item \textbf{Middle Finger}: Middle Finger MCP: $(L_9, L_{10}, L_{11})$; Middle Finger PIP: $(L_{10}, L_{11}, L_{12})$
    \item \textbf{Ring Finger}: Ring Finger MCP: $(L_{13}, L_{14}, L_{15})$; Ring Finger PIP: $(L_{14}, L_{15}, L_{16})$
    \item \textbf{Pinky Finger}: Pinky MCP: $(L_{17}, L_{18}, L_{19})$; Pinky PIP: $(L_{18}, L_{19}, L_{20})$
\end{itemize}
The corresponding full forms of the names of the joints "CMC", "MCP", "PIP" have been mentioned in detail in Mediapipe \cite{lugaresi2019mediapipeframeworkbuildingperception} documentation.

Here $L_i$ corresponds to the $i$th node as depicted in Figure \ref{fig:Landmarks}.

The angles were calculated by identifying three key points (landmarks) on each finger and applying vector mathematics to determine the angles at specific joints. The vectors between these points are calculated, and the angle between the vectors is determined using the dot product formula:

\begin{equation}
\text{Angle} = \arccos\left(\frac{\mathbf{u} \cdot \mathbf{v}}{|\mathbf{u}||\mathbf{v}|}\right)
\end{equation}

where $\mathbf{u}$ and $\mathbf{v}$ are the vectors formed by the landmarks.

The output of this step is an undirected spatial graph $G(V, E)$ with the set $V$ containing the 21 nodes (or, vertices) corresponding to the 21 extracted landmarks and the edge set $E$ consists of the undirected edges connecting the nodes.

\subsubsection{Normalizations}     
\textbf{1. Translational Normalization}:  
Translational normalization on graphs is performed to center the data around the origin, which helps in reducing the variance due to positional differences. This normalization improves the consistency and comparability of graph-based features.  
 This is achieved by subtracting the coordinate of a reference node from each node's coordinates. 
\begin{equation}
\mathbf{x}_i' = \mathbf{x}_i - \mathbf{x}_0
\end{equation}
Where $\mathbf{x}_i'$, $\mathbf{x}_i$, $\mathbf{x}_0$ are the normalized coordinate of the $i$th node, original coordinate of the $i$-th node and coordinate of the reference node (which is the wrist at position 0 as depicted in Figure \ref{fig:Landmarks}) respectively.

\vspace{0.1in}
\textbf{2. Scale Normalization}:  
To standardize the input data, we scaled the hand landmarks such that the maximum distance between any two landmarks is equal to a fixed desired distance. This normalization improves consistency and comparability across different samples.

Given a set of hand landmarks $\mathbf{X} = \{\mathbf{x}_i\}_{i=1}^N$ where $\mathbf{x}_i \in \mathbb{R}^3$, the scaling process is as follows:

We compute all pairwise Euclidean distances between the 21 landmarks:
\begin{equation}
d_{ij} = \|\mathbf{x}_i - \mathbf{x}_j\|_2 \quad \forall \, i, j \in \{1, \ldots, N\}
\end{equation}

Next, we find the maximum distance in the current set of landmarks:

\begin{equation}
d_{\text{max}} = \max_{i,j} (d_{ij})
\end{equation}

Following this, we calculate the scale factor as the ratio of the desired maximum distance $d_{\text{desired}}$ to the current maximum distance $d_{\text{max}}$:
\begin{equation}
s = \frac{d_{\text{desired}}}{d_{\text{max}}}
\end{equation}

Finally, scale all the landmarks by this scale factor:
\begin{equation}
\mathbf{x}_i' = s \cdot \mathbf{x}_i \quad \forall \, i \in \{1, \ldots, N\}
\end{equation}

\subsubsection{Data creation}
The data, from each graph $G(V, E)$, is finally represented as an adjacency matrix and a node embeddings matrix. The degree matrix $D$ calculated as a diagonal matrix containig the number of edges attached to each graph node, was calculated later during training of the architecture.\\
\textbf{Adjacency Matrix}: An adjacency matrix is a square matrix used to represent a finite graph, with elements indicating whether pairs of vertices are adjacent or not. If there is an edge between vertices $i$ and $j$, the matrix entry $A[i][j]$ is set to 1; otherwise, it is set to 0. It was calculated by iterating through the edges of the graph and marking the corresponding entries in the matrix. The final shape of the matrix for each graph was $(21 \times 21)$ \\
\textbf{Node embedding matrix}: For each node in the graph, its embeddings were the coordinates $(x, y, z)$ and its joint angle if available. If the joint angle is not available, that entry is marked as 0. The final shape of the matrix for each graph was $(21 \times 4)$.

\begin{figure}[!h]
    \centering
    \includegraphics[width=0.4\linewidth]{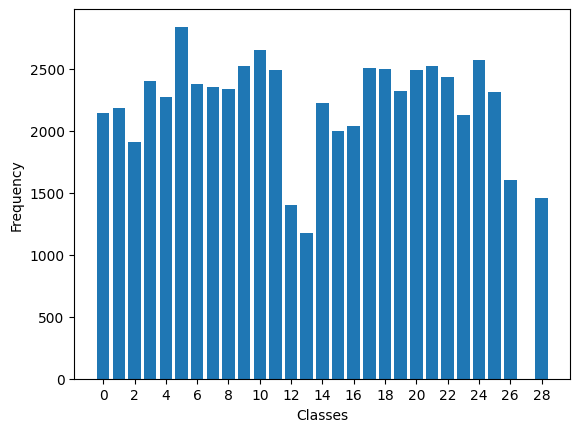}
    \caption{Class distribution of the number of data instances for each class. It is to be noted that only 1 instance of data could be extracted for the class 27, i.e, for the class "DELETE".}
    \label{fig:enter-label}
\end{figure}

\begin{figure*}[h]
    \centering
    \includegraphics[width=\textwidth]{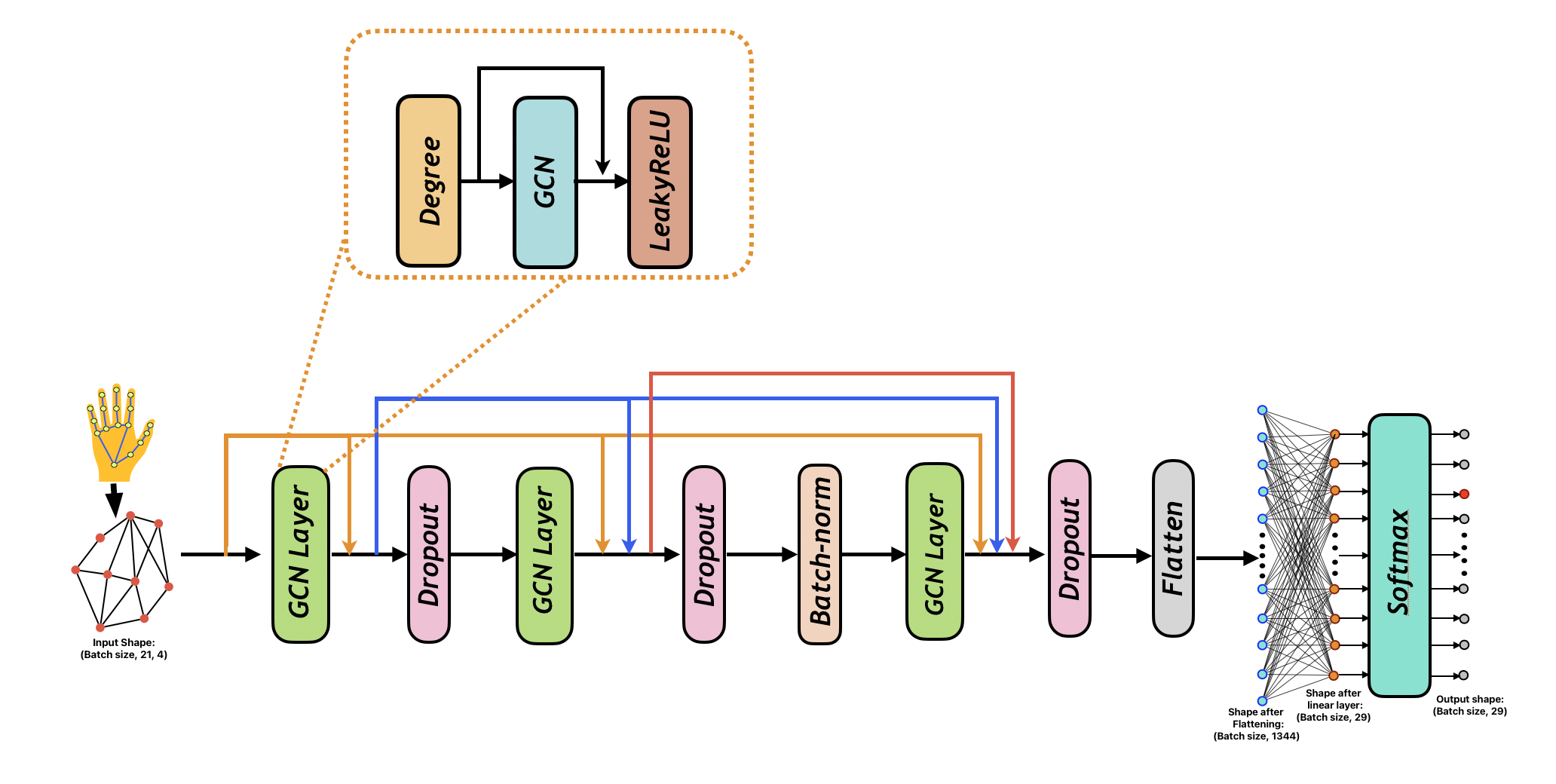}
    \caption{Network Architecture showing GCN Layers and Successive Residual connections. Total number of parameters in the model: 142447.}
    \label{fig:architecture}
\end{figure*}

\subsection{Network Architecture}
Please refer to Figure \ref{fig:architecture}. The proposed architecture consists of 3 "GCN" layers having Dropout Layers attached with them. \textbf{Successive Residual connections} \cite{he2015deepresiduallearningimage} for each layer were implemented to mitigate the effect of \textbf{shattered gradients} \cite{balduzzi2018shatteredgradientsproblemresnets}. Hence the input was added to the outputs of first, second and third GCN layers; output of first GCN layer was added to outputs of second and third GCN layers; output of second GCN layer was added to the output of the third GCN layer. The GCN layer itself has a residual connection in it. These create a \textbf{smoother error surfaces} \cite{li2018visualizinglosslandscapeneural}. Including residual connections also creates multiple networks acting in parallel and these include networks with fewer layers. A single batch normalization layer was included after the second GCN layer to reduce the issue of vanishing and exploding gradients and combat \textbf{internal covariate shift} \cite{10.5555/3045118.3045167}. 
The final output after the GCN layers were flattened into a one-dimensional vector and passed through a single layer feed forward network to ouput a vectors of shape $(\textit{batch size}\times 1 \times 29)$. This vector was then passed through the Softmax function for classification purpose.

\subsubsection{GCN Layer}
The GCN layer incorporates 2 basic functions of a neural message passing Graph Neural Network: (1) \textbf{Aggregation step:} For each node $n$ in the graph, messages are passed to that node from its neighbours. Here, the neighnours of a node $n$ are those nodes having a step distance of 1 from the node of interest. 
The input embeddings of a layer are given by $\mathbf{h}_{\text{in}}$. The aggregation step is $$\mathbf{Z}_{out} = \mathbf{h}_{\text{in}} \mathbf{W}$$ where $\mathbf{W}$ is the shared weight matrix for node-wise feature transformation. \\
(2) \textbf{Update step:} The update step updates the embeddings of every node after the Aggregation step. The update step is given by the output of the aggregation step:
\[
\mathbf{h}_{\text{out}} = D^{-\frac{1}{2}} (A + I) D^{-\frac{1}{2}} \mathbf{Z}_{\text{out}}
\]
where \( D \) is the degree matrix, \( A \) is the adjacency matrix which we augment with self-loops representing self-connections for every node, and \( I \) is the identity matrix.  The Degree matrix $D$ is calculated at the initial step of the GCN layer.   
Finally, the output of the GCN layer is passed through the LeakyReLU activation function before connecting it to the input of the GCN layer through a residual connection.

\section{Experiments}
\subsection{Training Details}
The hyperparameters chosen for our experiments were obtained using the hyperparameter optimisation package \textbf{Optuna} 
\cite{akiba2019optunanextgenerationhyperparameteroptimization}
The hyperparameters finally chosen have been reported in Table \ref{tab:hyperparameters}.

\captionsetup{justification=centering}
\begin{table}[htbp]
    \centering
    \captionsetup{width=.8\textwidth}
    \caption{Hyperparameter Values}
    \begin{tabular}{ll}
        \toprule
        \textbf{Hyperparameter} & \textbf{Value} \\
        \midrule
        Learning Rate      & 3 $\times 10^{-4}$ \\
        Batch Size         & 64 \\
        Dropout Probability       & 0.4 \\
        Adam Weight Decay       & $1 \times 10^{-4}$ \\
        $\beta_1$       & 0.9 \\
        $\beta_2$       & 0.999 \\
        LeakyReLU leak $\alpha$  & 0.2 \\
        EarlyStopping Callback patience & 15\\
        \bottomrule
        \label{tab:hyperparameters}
    \end{tabular}
\end{table}

We empirically observed that a mini-batch size less than 64 led to unstable validation metrics. 
We decided to proceed with a more complicated architecture and employ regularization techniques to capture more complex patterns in the data.
The \textbf{dropout layers} \cite{JMLR:v15:srivastava14a} as shown in Figure \ref{fig:architecture} were made by dropping a subset of graph nodes and edges randomly with a particular probability.
The \textbf{Adam optimiser} \cite{kingma2017adammethodstochasticoptimization} was used with moment estimate parameters $\beta_1$ and $\beta_2$. Weight decay was also employed on the optimizer as a regularization technique.

The \textbf{negative log-likelihood loss} (\texttt{NLLLoss}) was used as the loss function. It measures the performance of a classification model whose output is a probability distribution over classes. The formula for \texttt{NLLLoss} is:

\[
\text{NLLLoss} = -\frac{1}{N} \sum_{i=1}^{N} \log(P(y_i)),
\]

where \(P(y_i)\) is the predicted probability for the true class \(y_i\) and \(N\) is the number of samples(here, it was 29 for the 29 classes of the dataset).  \\
We used an \textbf{Early stopping callback} in our training phase with a pre-defined arbitrary number of epochs as patience, which means that the training was terminated if there was no improvement in the validation loss after that many epochs of training. \\
Only the weights of the epoch showing the best validation loss was used as the final model weights. \\
The model weights were initialized using \textbf{Xavier initialization} \cite{glorot2010understanding}, also known as Glorot initialization. This initialization helps in preventing the vanishing or exploding gradient problems, leading to more stable and faster convergence during training. \\
The model training was done using the PyTorch framework on Nvidia Tesla P100 GPU with a 16GB RAM.



\subsection{Results and Discussion} 
  
  To rigorously evaluate the performance of our proposed architecture, we employed a 5-fold cross validation process using the metrics \textbf{accuracy, recall, precision, F1-score, Area under the Curce (AUC) score, number of training epochs} and \textbf{Average time per epoch}.

\begin{table*}
\centering
\caption{Validation Accuracies of Various Models on ASL Fingerspelling A Dataset. It is to be noted that only our proposed architecture was trained by us on the ASL dataset. The metrics for the other architectures have been reported from literature. The accuracy of our architecture was reported as the average of the accuracy values found in the 5 folds of the cross validation process. The best performing accuracy has been made bold.}
\label{accuracy_table}

\begin{tabular*}{\textwidth}{@{\extracolsep{\fill}} l c}
\toprule
\textbf{Model} & \textbf{Validation Accuracy} \\ 
\midrule
Max-pooling Convolutional Neural Networks\cite{nagi2011max} & 88\% \\ 
Real-Time ASL Fingerspelling Recognition\cite{pugeault2011spelling} & 92\% \\ 
Real-Time Sign Language Recognition Using Depth Camera\cite{kuznetsova2013real} & 90\% \\ 
ASL Alphabet Recognition Using Microsoft Kinect\cite{dong2015american} & 93\% \\ 
Multi-Modal Deep Hand Sign Language Recognition\cite{rastgoo2018multi} & 95\% \\ 
Real-Time Sign Language Fingerspelling Recognition using CNN\cite{kang2015real} & 94\% \\ 
Our Architecture & \textbf{99.14\%} \\ 
\bottomrule
\end{tabular*}
\end{table*}

\begin{table*}[h!]
\centering
\caption{Model Performance Metrics on Training and Testing data with 5 fold cross validation. The best testing metrics have been made bold.}
\begin{tabular*}{\textwidth}{@{\extracolsep{\fill}} lcccccc}
\toprule
Type & Accuracy & Precision & Recall & F1 Score & Macro AUC & Weighted AUC \\
\midrule
Testing data  & 0.992 & 0.991 & 0.992 & 0.991 & 0.990 & 0.990 \\
Training data & 0.991 & 0.990 & 0.990 & 0.990 & 0.991 & 0.991 \\
\midrule
Testing data  & 0.990 & 0.992 & 0.991 & 0.992 & 0.991 & 0.991 \\
Training data & 0.991 & 0.991 & 0.993 & 0.991 & 0.992 & 0.992 \\
\midrule
Testing data  & 0.993 & 0.994 & 0.992 & 0.994 & 0.993 & 0.993 \\
Training data & 0.990 & 0.993 & 0.992 & 0.993 & 0.994 & 0.994 \\
\midrule
Testing data  & 0.989 & 0.993 & \textbf{0.994} & 0.993 & 0.992 & 0.992 \\
Training data & 0.992 & 0.992 & 0.991 & 0.992 & 0.993 & 0.993 \\
\midrule
Testing data  & \textbf{0.994} & \textbf{0.995} & 0.992 & \textbf{0.995} & \textbf{0.994} & \textbf{0.994} \\
Training data & 0.990 & 0.994 & 0.994 & 0.994 & 0.995 & 0.995 \\
\bottomrule
\end{tabular*}
\label{cross-val-metrics}
\end{table*}

\begin{figure}[H]
    \centering
    \includegraphics[width=0.4\linewidth]{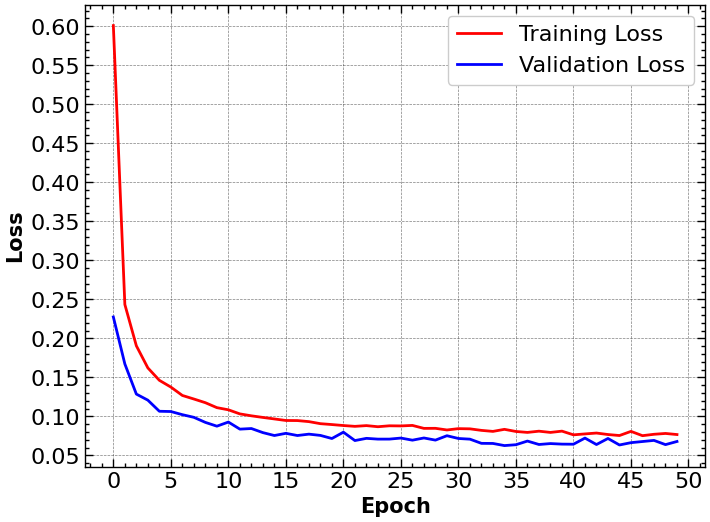}
    \caption{Training and validation losses measured during training. The validation loss curve is always below the training loss curve because of the much fewer number of data instances it was calculated upon. There is a slight instability in the validation loss curve which is due to the smaller batch-size taken during training. These loss curves correspond to the best performing fold on the validation loss in 5-fold cross validation.}
    \label{fig:Loss-curves}
\end{figure}

\begin{figure}[H]
    \includegraphics[width=1.0\linewidth]{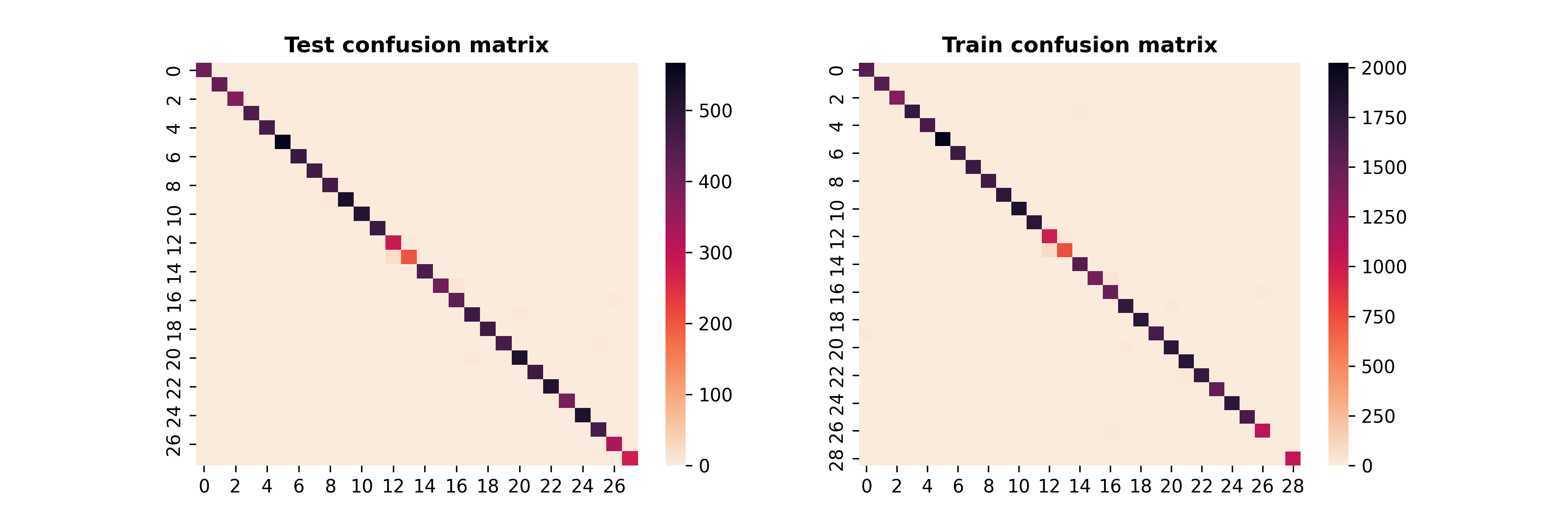}
    \caption{Training and testing confusion matrix. The overall diagonal structure of the model shows commendable performance of our model on the testing dataset.}
    \label{fig:conf-matrix}
\end{figure}

\subsubsection{Validation Metrics}
In our study, we compared our results with four other models previously published in the literature on the same dataset. These have been referred in \cite{nagi2011max}, \cite{pugeault2011spelling}, \cite{kuznetsova2013real}, 
\cite{dong2015american}, 
\cite{rastgoo2018multi}, 
\cite{kang2015real}. 
The accuracy metrics for the various architectures have been presented in Table \ref{accuracy_table}. \\
The loss curves during training and validation phase of our architecture have been plotted in Fig \ref{fig:Loss-curves}. The non-diverging nature of the loss curve shows that our model did not overfit on the dataset which is a common issue  \cite{towardsdatascience2020} on the ASL dataset.

Table \ref{cross-val-metrics} shows the results of our 5-fold cross validation strategy. Our model consistently performs above the 99\% threshold across various evaluation metrics, including accuracy, precision, recall, and AUC, on the validation sets during the 5-fold cross-validation process. This demonstrates the model's generalizability on the validation data.

The confusion matrix on the training and testing dataset have been presented in Figure \ref{fig:conf-matrix}. The general diagonal structure of the matrix shows that our model generalizes well on the testing data.

\subsubsection{Training Time}
The training time of our model was also calculated. Due to the Early Stopping callback used in our training loop, it is judicious to report the average training time per epoch and the total number of epochs, rather than the total training time of our model. These results have been presented in Table \ref{tab:training-time}.

\begin{table}[h!]
\caption{Average Training Time per Epoch and Number of Training Epochs. The model practically converged after about 30 epochs in each fold of training. The lowest average time per epoch has been made bold.}
\centering
\begin{tabular}{cc}
\toprule
\textbf{Average Time per Epoch (seconds)} & \textbf{Total Number of Epochs} \\
\midrule
12.5 & 50 \\
14.2 & 45 \\
\textbf{11.8} & 47 \\
13.3 & 48 \\
15.0 & 46 \\
\bottomrule
\end{tabular}
\label{tab:training-time}
\end{table}

\section{Conclusion}
In this study, we have introduced a novel framework for American Sign Language (ASL) recognition utilizing Graph Convolutional Networks (GCNs) integrated with successive residual connections. By leveraging the natural graph-based representation of hand landmarks, our approach effectively addresses the limitations of traditional Convolutional Neural Networks (CNNs), which are constrained by their grid-based data processing.

The proposed method and model enhance the ability to model the complex spatial relationships inherent in ASL gestures, leading to quite satisfactory results on the ASL dataset. 

 Future research is going upon this foundation by exploring additional graph-based representations, incorporating more complex network architectures, and applying the approach to larger and more diverse generalised sign language hand-shape datasets. 
 Any future datasets emerging out of new developments may be bench-marked by this model performance for testing on all applicable figures of merit.
 The findings presented in this paper contribute to the advancement of human-computer interaction technologies, particularly in the domain of assistive systems for the deaf and speech impaired communities.

\bibliographystyle{ieeetr}
\bibliography{main}

\end{document}